\documentclass[letterpaper]{article}
\usepackage{aaai}

\usepackage{times}

\usepackage{amssymb}
\usepackage{tikz,pgf}
\usetikzlibrary{positioning}
\usetikzlibrary{shapes.geometric} 
\usetikzlibrary{shapes}
\usepackage{pgfplots}
\usepackage{colonequals}

\newcommand{\pr}[1]{\mathbb{P}\!\left(#1\right)}

\newcommand{\bigperiod}{\mbox{\large\bfseries .}}

\begin{document}

\title{Speeding-up ProbLog's Parameter Learning}
\author{Francisco H. O. V. de Faria$^1$, Arthur C. Gusm\~ao$^1$, Fabio G. Cozman$^1$, Denis D. Mau\'a$^2$ \\
$^1$ Escola Polit\'ecnica, $^2$ Instituto de Matem\'atica e Estat{\'\i}stica \\
Universidade de S\~ao Paulo, S\~ao Paulo, Brazil}

\maketitle

\begin{abstract}
ProbLog is a state-of-art combination of logic programming and
probabilities; in particular ProbLog  offers parameter
learning through a variant of the EM algorithm. However, the resulting
learning algorithm is rather slow, even when the data are complete. In
this short paper we offer some insights that lead to orders of magnitude
improvements in ProbLog's parameter learning speed with complete data.
\end{abstract}

\section{Introduction}

There are many ways to combine logical expressions and probabilities
\cite{Getoor2007Book,Raedt2010}.
Sato's distribution semantics is perhaps the most popular proposal in the
literature \cite{Sato2001}. There one has rules as in   logic programming, for instance
\begin{equation}
\label{equation:Calls}
\mathsf{calls}(X,Y) \colonminus \mathsf{alarm}(X), \mathsf{neighbor}(X,Y)\bigperiod
\end{equation}
and probabilities such as $\pr{\mathsf{neighbor}(X,Y)}\!=\!0.3$, meaning that
the probability that any $X$ and $Y$ are neighbors is $0.3$.

ProbLog is a freely available package that allows one to manipulate and learn such
probabilistic logic programs  \cite{Fierens2015}.
The package is very friendly; one can easily use
it to represent knowledge about deterministic and probabilistic statements.
To do parameter learning (that is, to learn probability values for a given program),
ProbLog implements a variant of the EM algorithm that resorts  to BDD
diagrams so as to speed inference whenever possible \cite{Gutmann2011,Fierens2015}.

However, the EM-variant used by ProbLog is rather slow, {\em even} when there
is no missing data in the input. This is somewhat perplexing for a new user, who
may be surprised to find that even small propositional programs demand high
computation effort with complete data. In this short paper we analyze this
behavior of the EM-variant in ProbLog, and we detect the main reasons for
its inefficiency. We propose   techniques that
lead to orders of magnitude improvements in speed. We demonstrate these gains with experiments.
Even though the ideas in this short paper are in essence simple, they will be
important in bringing probabilistic logic programming to real applications.

\section{A very short review}

We consider the following syntax, entirely taken from the ProbLog package
as described by \citeauthor{Fierens2015} \shortcite{Fierens2015}.
A rule is written as $h \colonminus b_1, \dots, b_n\bigperiod$,
where $h$ is an atom, called the {\em head}, and each $b_i$ is an atom
perhaps preceded by $\mathbf{not}$. Each $b_i$ is a {\em subgoal} and the
left hand side is the {\em body}.
A rule without a body,
written $h\bigperiod$, is a {\em fact}. Rules and facts can be grounded by
replacing logical variables by constants.
The {\em dependency graph} of a program
is a graph where the nodes are the grounded atoms, and where there
is an edge from each grounded subgoal to the corresponding grounded head.
A program is {\em acyclic} if its dependency graph is acyclic.

A probabilistic fact, denoted by $\alpha::h\bigperiod$, consists of a number
$\alpha$, here assumed to be a rational in $[0,1]$, and an atom $h$.
A probabilistic fact may contain logical variables, in which case it is interpreted
as the set of grounded probabilistic facts produced by replacing logical variables
by constants in every possible way. Additionally, ProbLog allow for
{\em probabilistic rules}, written as $\theta::h \colonminus b_1,\dots, b_n\bigperiod$,
and interpreted as a pair consisting of a probabilistic fact $\theta::x\bigperiod$ and
a rule $h \colonminus b_1,\dots,b_n, x\bigperiod$, where $x$ is an auxiliary atom
(with the same logical variables as $h$)
that is not present anywhere else in the program.

Suppose we have a set of probabilistic rules/facts, but we do not
know the values of the probabilities. We can use a dataset $D$ to {\em learn those
parameters}; we assume this is done by choosing parameters
$\Theta$ that attain $\max_{\Theta} L(\Theta)$, where the {\em log-likelihood}
$L(\Theta)$  is the probability $\log \pr{D}$ with respect to parameters $\Theta$.
When $D$ has some missing data,
one popular way to maximize log-likelihood is to resort of the EM algorithm: here
one iterates between inference and
maximization of the expected log-likelihood. EM typically requires
computing the probability of each random variable together with the missing
variables that affect it (that is, the variable and its ``parents'') \cite{Darwiche2009}.

Parameter learning is done by ProbLog as follows. Any probabilistic rule is
written as a pair consisting of a fresh auxiliary probabilistic fact and a deterministic rule.
This guarantees that every probability is associated with an atom
that has no parents in the dependency graph: both the inference and the
maximization steps then become rather elementary \cite{Fierens2015}.

\section{A better algorithm}

A point to note is that, {\em even when the data are complete, the auxiliary facts introduced
to handle probabilistic rules are missing}. Thus ProbLog must run the EM-style algorithm {\em even}
when the input $D$ is complete. One can see the consequences of this
 in Figure \ref{figure:Time}:
{\em even} for small datasets, {\em even}
for propositional acyclic programs, learning takes too long.

Our solution is {\em not} to insert an auxiliary (latent) atom for each probabilistic rule.
Instead, we must write down the log-likelihood  and maximize it directly; the main
insight is that, for many rule patterns, this maximization can be done in closed-form.
Consider an example. Suppose we have two propositional rules with the same head,
say
\[
\theta_1::h\bigperiod \quad \mbox{and} \quad \theta_2::h \colonminus b\bigperiod,
\]
and a complete dataset with $N$ observations of $(h,b)$. The log-likelihood (restricted to this
head atom) is
\[
\begin{array}{c}
N_{00} \log (1-\theta_1) +
N_{01} \log (1-\theta_1-\theta_2+\theta_1\theta_2) + \\
N_{10} \log \theta_1 +
N_{11} \log (\theta_1+\theta_2-\theta_1\theta_2),
\end{array}
\]
where $N_{ij}$ is the number of times the configuration $\{h=i,b=j\}$
(taking $1$ to mean $\mathsf{true}$ and $0$ to mean $\mathsf{false}$).
This is apparently much more complex than the usual likelihood one finds
for example when learning Bayesian networks \cite{Darwiche2009}; however,
with some effort we find that the estimates that maximize log-likelihood are
\[
\hat{\theta}_1 = \frac{N_{10}}{N_{00}+N_{10}},
\qquad
\hat{\theta}_2 = \frac{N_{00} N_{11} - N_{10} N_{01}}{N_{00} N_{11} + N_{00} N_{01}}.
\]
In fact, a very large number of rule patterns admit similar exact solutions. In this implementation we have covered all possible patterns for combinations of at most
three rules entailing a same predicate. Notice that the log-likelihood expression does
not depend on the size of the rules' bodies. And whenever the log-likelihood (for the
rules that share the same head) does not admit a closed-form solution, we have found
that a fast gradient-based algorithm can quickly find its maximum.

Thus our modified learning algorithm (for complete data) is to
maximize likelihood directly, locally maximizing it in closed-form
whenever possible, or locally maximizing it numerically with
a fast gradient-based routine whenever necessary.

\section{Experiments}

We have implemented the techniques discussed in the previous section, by modifying
ProbLog's parameter learning code. To demonstrate that the techniques are indeed
effective, we present here two experiments; they are necessarily small because the
original ProbLog algorithm cannot handle large models, and we want to compare our
results with that previous algorithm. All of our tests were run in identical processors
at Amazon Web Services.

\begin{figure}
\hspace*{-2ex}
  \begin{tikzpicture}[scale=1.02]
    \begin{axis}[xmode=log,ymode=log,width=8.5cm,height=6cm,
      xlabel={Size of dataset}, ylabel={Time to finish (seconds)},
      legend style={at={(0.5,0.65)},nodes={scale=0.7, transform shape}, anchor=south west,legend columns=1},
      ]

      \addplot[black,mark=o,dashed,mark options={solid}]
      coordinates {
(5,148.1756668091)
(10,528.4635727406)
(15,354.4418897629)
(20,1714.26334095)
(25,2535.7322692871)
(30,3028.2452924252)
(35,2877.0464596748)
(40,3536.4865977764)
(45,3114.3571631909)
(50,5498.0638287068)
(55,6920.0679392815)
      };
      \addlegendentry{Propositional, original}

      \addplot[blue,mark=*]
      coordinates {
(5,0.812130212783813)
(10,1.07050228118896)
(15,1.05292868614196)
(20,1.33560371398925)
(25,1.00481724739074)
(30,1.68554615974426)
(35,1.51514506340026)
(40,1.88365864753723)
(45,1.85300087928771)
(50,2.01372075080871)
(55,1.9037492275238)
(60,1.79057788848876)
(65,1.72767996788024)
(70,2.08643412590026)
(75,2.17253875732421)
(80,2.50885319709777)
(85,2.00534057617187)
(90,2.32351613044738)
(95,2.12786102294921)
(100,2.59070777893066)
(150,2.39665079116821)
(200,3.3414671421051)
(250,3.39863872528076)
(300,4.52638435363769)
(350,4.20475149154663)
(400,4.34412217140197)
(450,5.43639969825744)
(500,5.61913537979126)
(550,6.06050562858581)
(600,6.18991732597351)
(650,7.11157274246215)
(700,7.07214593887329)
(750,6.91987419128418)
(800,7.20532393455505)
(850,8.19364070892334)
(900,7.79710817337036)
(950,8.2199535369873)
(1000,9.2636706829071)
      };
      \addlegendentry{Propositional, proposed}

      \addplot[black,mark=square,dashed,mark options={solid}] coordinates
      {
(5,3.5548355579)
(10,45.1929433346)
(15,179.6756141186)
(20,569.4401385784)
(25,1863.0277516842)
      }; \addlegendentry{Relational, original}

      \addplot[blue,mark=square*]
      coordinates{
(5,0.113916397094726)
(10,0.180948257446289)
(15,0.381650447845459)
(20,0.557823181152343)
(25,0.700496673583984)
(30,0.977568864822387)
(35,1.20295405387878)
(40,1.57065486907958)
(45,1.85382318496704)
(50,2.12604713439941)
(55,2.527836561203)
(60,3.09825801849365)
(65,3.57150316238403)
(70,4.08850502967834)
(75,4.74174070358276)
(80,5.29474663734436)
(85,5.99687790870666)
(90,6.4289538860321)
(95,7.37290263175964)
(100,7.931396484375)
(105,8.90825247764587)
(110,9.66104078292846)
(115,10.5723400115966)
(120,11.6514480113983)
      };
    \addlegendentry{Relational, proposed}

      \addlegendentry{AMPU}

    \end{axis}
  \end{tikzpicture}
  \caption{Time to learn parameters from data (note log-scale!). In the propositional case, size of dataset is the number of observations for all atoms;
  in the relational case, size of dataset is the the number of constants in the program. }
  \label{figure:Time}
\end{figure}
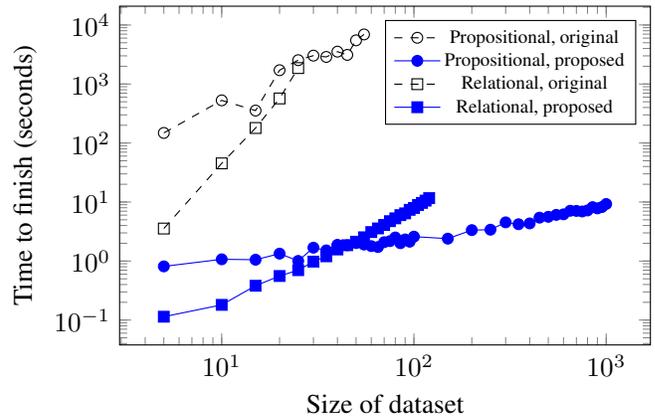

So, consider first an acyclic propositional program that encodes the energy plant of a ship\footnote{We
have obtained the model from the site http://www.machineryspaces.com/emergency-power-supply.html;
this is an ``almost'' deterministic system in the sense that several relations are Boolean, together
with sources of random noise.}
 using 16  propositions, 17  probabilistic rules and  7  probabilistic facts.
This is a relatively small program, yet the original ProbLog algorithm requires significant
computer time, as can be seen in Figure~\ref{figure:Time}. We should note that
our algorithm found similar values for the log-likelihood in all cases; that is, by
introducing auxiliary atoms, ProbLog makes the maximization harder without making it more effective.
For a propositional dataset with 50 observations, ProbLog reaches a log-likelihood of $-176.65$
in $5498.06$ seconds, while our algorithm reaches a log-likelihood of $-176.60$ in $2.02$ seconds.

%
%
%
%
%
%

The second example is a short relational program consisting of:
\[
\begin{array}{c}
\theta_1::\mathsf{fire}(X)\bigperiod, \\
\theta_2::\mathsf{burglary}(X)\bigperiod, \\
\theta_3::\mathsf{neighbor}(X,Y)\bigperiod,  \\
\theta_4::\mathsf{alarm}(X) \colonminus \mathsf{fire}(X)\bigperiod, \\
\theta_5::\mathsf{alarm}(X) \colonminus \mathsf{burglary}(X)\bigperiod,\\
\theta_6::\mathsf{calls}(X,Y) \colonminus \mathsf{neighbor}(X,Y),\mathsf{alarm}(Y)\bigperiod.
\end{array}
\]
Suppose we have   $N$ constants, each one of them denoting a person
in some city. Figure \ref{figure:Time} compares the computational effort spent
by the original ProbLog and our algorithm ($N$   corresponds to
the dataset size in the propositional case). Our algorithm outperforms the
original ProbLog algorithm in computer time, reaching similar log-likelihood values. For a
relational dataset with 25 constants, ProbLog reaches a log-likelihood of $-523.11$ in $1863.03$ seconds,
while our algorithm reaches a log-likelihood of $-523.11$ in $0.69$ seconds.

\section{Conclusion}

We have shown that ProbLog's parameter learning algorithm can be dramatically
improved  by resorting to a few insights:
first, never introduce unnecessary  latent atoms; second, maximize
  log-likelihood locally in the most efficient manner (closed-form or gradient-based).
These insights will be helpful in future work dealing with missing data and with rule learning.

\section{Acknowledgement}
The first author is supported by a scholarship from Toshiba Corporation.
The second author is supported by a scholarship from CNPq.
The third and fourth authors are partially supported by CNPq.
This work was partly supported by the S\~ao Paulo Research Foundation (FAPESP) grant 2016/01055-1 and the CNPq grants 303920/2016-5 and 420669/2016-7; also by S\~ao Paulo Research Foundation (FAPESP) grant 2016/18841-0 and CNPq grant 308433/2014-9; finally by FAPESP 2015/21880-4.
\bibliographystyle{aaai}
\bibliography{consulted}
\end{document}